\documentclass[showpacs,aps,pra,twocolumn,superscriptaddress,longbibliography,nofootinbib]{revtex4-2}
\usepackage{geometry}
\usepackage{amsfonts}
\usepackage{amssymb}
\usepackage{pgfplots}
\pgfplotsset{compat=1.18}
\usepackage{filecontents}
\usepackage{graphicx}
\usepackage{hyperref}

\usepackage{breakcites}
\usepackage{amsmath}
\usepackage{hyperref}
\usepackage{cleveref}
\usepackage{mathtools}
\usepackage{physics}
\usepackage{tikz}     
\usepackage{graphicx}
\usepackage{color}
\usepackage{microtype}
\usepackage[utf8]{inputenc}
\usepackage{comment}
\usepackage{tcolorbox}
\usepackage{ragged2e}
\usepackage{xcolor}
\usepackage{soul}
\usepackage{booktabs}

\usetikzlibrary{positioning}
\usetikzlibrary{calc}
\usetikzlibrary{arrows.meta}

\colorlet{lightblue}{blue!15}
\colorlet{lightgreen}{green!15}
\colorlet{lightred}{red!15}
\colorlet{lightyellow}{yellow!10!white}
\colorlet{darkblue}{blue!80!black}
\definecolor{applegreen}{rgb}{0.55,
0.71, 0.0}
\definecolor{amber}{rgb}{1.0,
0.49, 0.07}

\begin{document}
\title{How good is my story? Towards quantitative metrics for evaluating LLM-generated XAI narratives}

\author{Timour Ichmoukhamedov}
\email{timour.ichmoukhamedov@uantwerpen.be}
\affiliation{ADM, Universiteit Antwerpen,  Prinsstraat 13 , 2000 Antwerp, Belgium}
\author{James Hinns}
\email{james.hinns@uantwerpen.be}
\affiliation{ADM, Universiteit Antwerpen,  Prinsstraat 13 , 2000 Antwerp, Belgium}
\author{David Martens}
\affiliation{ADM, Universiteit Antwerpen,  Prinsstraat 13 , 2000 Antwerp, Belgium}
\date{\today}

\begin{abstract}
A rapidly developing application of LLMs in XAI is to convert quantitative explanations such as SHAP into user-friendly narratives to explain the decisions made by smaller prediction models. Evaluating the narratives without relying on human preference studies or surveys is becoming increasingly important in this field. In this work we propose a framework and explore several automated metrics to evaluate LLM-generated narratives for explanations of tabular classification tasks. We apply our approach to compare several state-of-the-art LLMs across different datasets and prompt types. As a demonstration of their utility, these metrics allow us to identify new challenges related to LLM hallucinations for XAI narratives.
\end{abstract}

\pacs{}

\maketitle

\section{INTRODUCTION}\label{sec:introduction}

\begin{figure*}[!htbp]
\includegraphics[width=0.95\textwidth, angle=0]{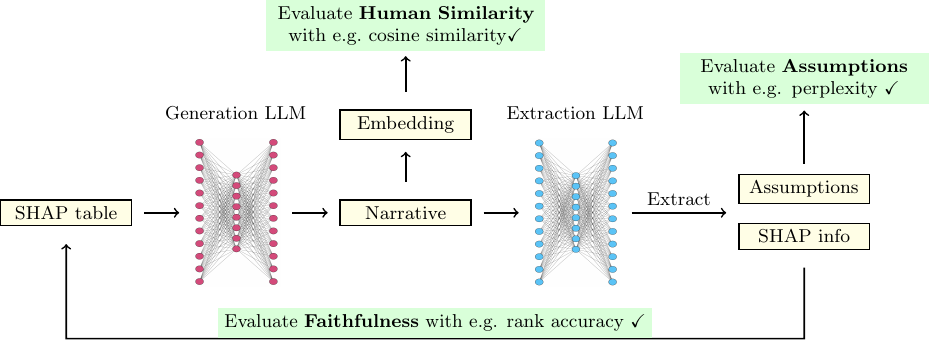}
\caption{Overview of the workflow presented in the paper. First, a narrative based on a SHAP input table is generated by the generation LLM, after which an extraction LLM is independently used to extract the pieces of information included into the narrative related to faithfulness and assumptions. The faithfulness to the original input table is then checked with downstream metrics such as rank or sign accuracy. For assumptions that were injected into the story by the LLM that cannot be checked against the original SHAP, downstream metrics like perplexity are explored. Finally, we explore an additional pathway for human similarity by embedding the generated narratives and comparing them to embeddings of human written narratives with metrics e.g.~cosine similarity.}
\label{fig:Schematic}
 \end{figure*}

One of the most popular feature attribution explanation tools for narrow black-box ML models in the field of Explainable AI (XAI) is SHAP \cite{Lundberg2017}. It combines ideas of using local linear approximations from LIME \cite{Tulio2016} with older game-theoretic Shapley values \cite{shapley1953} to estimate for every feature its relative effect on the model output after weighing it in all possible coalitions with the other features. A SHAP explanation for a particular instance therefore consists of a table (or other pictorial representations) of all features together with their contribution to the output probabilities relative to some base level. Regardless of its widespread use, it must be noted that SHAP is not without flaws and several recent criticisms can be found \cite{huang2023, silva2024}. More importantly, SHAP or other feature importance explanations can contain quite dense or rather subtle information consisting out of dozens of values that might not always be easily interpretable by non-experts. In the context of XAI this observation naturally gives raise to the question: \textit{Can the relative importance of multiple features reveal a richer or more easily interpretable story than the sum of its parts?} 

\textbf{Narratives:} This motivates the idea of complementing SHAP with a more laypeople oriented and user-friendly textual narrative that provides a plausible summary as to why and how the most important features contribute. Recent advances in Large Language Models (LLMs) open a novel opportunity to automate and scale this process without humans in the loop. Indeed, the idea of using LLM-generated narratives for XAI has been already explored in several recent works for tabular data \cite{Burton2023,martens2023,zytek2024} and similar developments are happening for graphs \cite{giorgi2024,pan2024,cedro2024} and images \cite{martens2023,Wojciechowski2024}. Initially, the task was viewed as a data-to-text task with the goal of strictly summarizing the quantitative XAI results to the user, and it has been shown that smaller models such as T5 or BART can be fine-tuned for this purpose \cite{Burton2023}. However, more recently, larger models such as GPT4 have been demonstrated to produce rich and fluent narratives that were found to be accessible to users \cite{martens2023}. In particular, the narratives generated in \cite{martens2023} go beyond simple data-to-text summaries and leverage the broader knowledge base of modern LLMs to add context and reasoning into the story. Similar results have also been obtained in a more recent pilot study in \cite{zytek2024}, where the study participants preferred narratives over SHAP-based plots on a range of metrics.

Interestingly, it has been also shown that even in the absence of a quantitative explanation such as SHAP, faithful narratives can still be generated by providing the LLM with carefully selected examples of model predictions \cite{Kroeger2024}. Other related work that also leverages LLMs on top of SHAP and LIME is TalkToModel \cite{Slack2023}, where the goal is to provide an interactive framework for a user to ask questions about the model.

\vspace{10pt}

\textbf{Evaluation:} Evaluating and scoring narratives is a challenging task and we do not aim to provide a fully exhaustive or universal framework. Two commonly used criteria for XAI explanations are \textit{faithfulness} and \textit{plausibility}\footnote{Not to be confused with the various definitions of plausibility in counterfactual explanations \cite{ijcai2021p609}.} \cite{Jacovi2020}. In short, faithfulness refers to the accuracy of an explanation relative to the inner workings of a model, and plausibility refers to how convincing the explanation is to a user. For the metrics explored further on, it will be convenient to make a division in the following categories:

\begin{itemize}
    \item \textbf{Faithfulness:} In the present work we will evaluate the faithfulness of a narrative relative to the explanation and other data provided to the LLM. In the assumption that the SHAP table is itself fully faithful, this also captures the faithfulness of the narrative relative to the prediction model.  
    \item \textbf{Human Similarity:} The narratives are similar to a set of reference narratives e.g.~written by human experts. This is a broad category that depending on the specific metrics might overlap with faithfulness, plausibility, or both.
    \item \textbf{Assumptions (Plausibility):} A narrative might have assumptions that contain general knowledge beyond what can be checked in the underlying explanation. For example, in the statement \textit{"The large number of fouls made by the team decrease the chances of getting an award because aggressive play is generally not appreciated in football"}, the assumption would be \textit{"Aggressive play is not appreciated in football"}. Assumptions fall outside of faithfulness and having reasonable assumptions is part of the plausibility.
\end{itemize}
It is important to emphasize that although these broad categories can be applied to any XAI narrative, one can imagine numerous metrics to measure them that might in addition also depend on the specific narrative style. For example, \textit{should a narrative provide the actual SHAP values, or simply convey order of importance?} To emphasize the fact that the metrics used in the final steps of our evaluation framework as depicted on Fig.~\ref{fig:Schematic} are not universal and supposed to be replaceable, we will often refer to them as downstream metrics. 

Existing XAI narrative research explores various directions for metrics, that we briefly discuss below. Perhaps the most obvious one is to perform user surveys such as in \cite{martens2023, zytek2024}, where the narratives are scored by humans on a broad range of questions. However, the main disadvantage of this approach is that it cannot be easily scaled or used in real-time for automated validation. In addition, existing user surveys on this topic do not explore the faithfulness of the narratives, but rather focus on user preferences. While human preference studies are of undeniable importance to obtain feedback from the final user, we will not consider them here and explore the direction of more automated metrics. 

Several other XAI narrative metrics (including user studies) can be found in the context of graphs in \cite{giorgi2024,pan2024} or images in \cite{Wojciechowski2024}. The most relevant and closest related recent work for the present paper is based on tabular data that explores both faithfulness and human similarity in similar ways as previously defined \cite{Burton2023}. However, for the faithfulness the authors perform a manual error analysis on a sample of narratives which once again indicates the relevance of more automated metrics. The human similarity on the other hand is evaluated using automated metrics such as BLEURT \cite{SELLAM2020} or METEOR \cite{Banerjee2005} (assuming that reference narratives are available). Although the use of an embedding based metric like BLEURT for narrative evaluation is a promising idea, it does not provide further insight into the reasons for semantic dissimilarity and should therefore be used only as a complementary metric. Was the narrative simply written in a less fluent way than human text or style, or is it making more fundamental errors? Moreover, at the moment of writing, BERT-based embeddings do not even fall in the top 100 of the MTEB embedding leaderboard\footnote{\url{https://huggingface.co/spaces/mteb/leaderboard}}. In light of this advance it would be interesting to explore the potential of more advanced embedding models as a way to evaluate narratives. Finally, the narratives generated in \cite{Burton2023} are summaries of the explanation, and do not contain any properties along the Assumption category introduced above.  

\begin{figure*}
    \includegraphics[width=0.95\textwidth, angle=0]{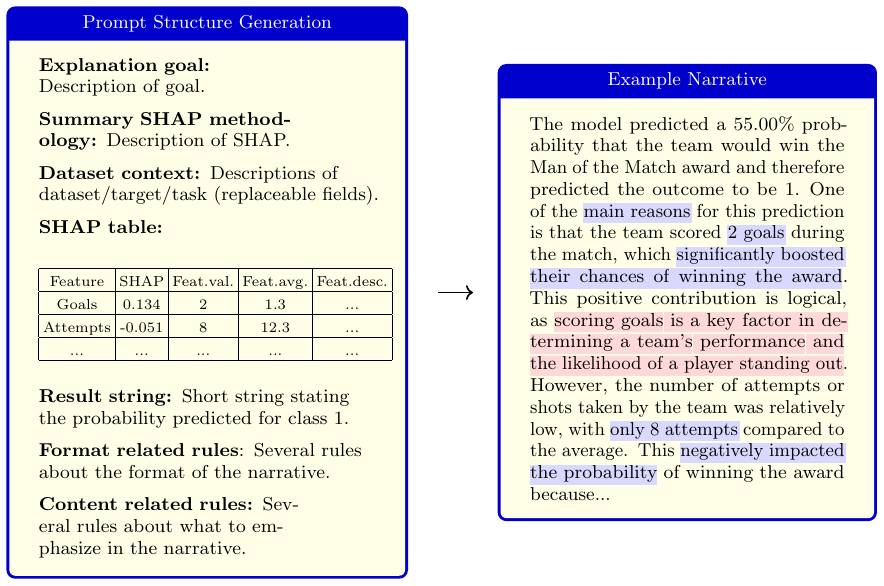}
    \caption{\textbf{Left:} A brief summary of the zero-shot prompt we use to generate the narratives. \textbf{Right:} Excerpt from a narrative generated with gpt-4o for the Fifa Man of the Match dataset. We highlight elements of the narrative that relate to faithfulness in blue and  the assumptions injected by the model in red.}
    \label{fig:GenerationPrompt}
\end{figure*}

\textbf{Present work:} In the present work we address the concerns raised above and make the following contributions to the state-of-the-art of narrative evaluation:

\begin{enumerate}
    \item Proposing an automated framework with multiple metrics to evaluate narratives across the categories introduced above (Faithfulness, Human Similarity, Assumptions).
    \item Exploring and validating the behavior of the metrics on several proof-of-concept experiments to establish their trustworthiness.
    \item Applying the metrics to compare narrative generation across several datasets and LLMs and demonstrate how they allow identifying new challenges in LLMs for XAI. 
\end{enumerate}

 A schematic overview of our methodology is presented in Fig.~\ref{fig:Schematic}. Most importantly, to achieve full automation we introduce an extraction model that can extract various quantities of interest in the narrative that could then be checked using downstream metrics of choice. The feasibility of this idea is backed by recent demonstrations that LLMs can achieve comparable performance to human annotators on various tasks \cite{MIN2023,Chen2023}. Automated extractions from a text as an intermediate step for error validation have also been used in a closely related problem of data-to-text generation \cite{REN2023}. The use of an extraction model could in addition lay groundwork for an application of the rapidly growing interest in agentic LLM approaches \cite{MASTERMAN2024}. In our approach we aim the extraction step towards faithfulness and assumptions as they relate to localized information in the narratives. 

 Finally, we also consider the scenario where a reference set of explanations together with expert-written narratives is available and can be used for evaluation. We demonstrate that even when relying on existing embedding-based metrics such as cosine similarity, we can in the majority of cases: (1) find the matching LLM-generated narrative to the expert narrative from a sample more reliably than BLEURT (2) detect a shift in similarity when the internal logic of the matched narrative is manipulated. This demonstrates that pretrained state-of-the-art embedding models can already (partially) measure narrative accuracy, and opens up the road towards specialized XAI narrative metrics beyond BLEURT. 

\section{Narrative generation}\label{sec:generation}

In what follows we briefly describe the narrative generation step and refer to the paper repository\footnote{\href{https://github.com/ADMAntwerp/SHAPnarrative-metrics}{https://github.com/ADMAntwerp/SHAPnarrative-metrics}} for more details and code. To generate the narratives we follow a conceptually similar approach to \cite{martens2023}, using a zero-shot prompt that briefly explains the task at hand, provides a description of the dataset, the features, and includes a SHAP table and prediction scores. We have explored passing either the full SHAP table, or a truncated table that only contains the most important features (in our case four) that need to be in the narrative, and decided to proceed with the latter for brevity. The truncated SHAP table is passed as a string with every row containing a feature name, the SHAP value, the feature value, the average feature value over the training set, and a short feature description. We will consider a feature ranking by absolute SHAP values, and hence immediately sort the truncated table in this order. In this work we restrict ourselves to the explanation of binary classification models with labels 0 and 1, but generalizations to multiple classes should be straightforward.

The structure of the prompt is illustrated in Fig.~\ref{fig:GenerationPrompt}. It is quite general and in principle allows to independently explore the effects of various content and format rules and optimize them towards user-preference. The optimal content and format related rules are not universal and likely to depend on the use case and target audience. This however falls beyond the scope of the present work, and we proceed with a single set of rules found to provide reasonable narratives through trial and error inspired from \cite{martens2023}. An example of a narrative generated in this way is also presented in Fig.~\ref{fig:GenerationPrompt}.

\textbf{Content related rules}: Our set of content rules are selected to avoid ambiguities related to faithfulness or assumptions and to clarify which aspects we would like to be emphasized. For this purpose we instruct the model to (1) emphasize to the reader the rank (in absolute value) and sign of a feature and (2) insert short additional suggestions as to why a feature could have contributed in this way. This second step leverages the broad general knowledge base of state-of-the-art LLMs and goes beyond previous work that was aimed at simply summarizing the SHAP table \cite{Burton2023}. To avoid confusion, we emphasize that in this context we generally assume the SHAP table is correct and faithfulness does not refer to the faithfulness of the explanation itself, but to the faithfulness of the narrative to the explanation.  While both faithfulness and assumptions (or their generalizations) can be expected to be present in any narrative based on an XAI explanation, the suitable metrics to measure them might vary. 

\textbf{Format related rules}: The main goal of this work is to focus on the information present in the narratives as divided in the categories introduced previously, and we will put a smaller emphasis on exploring the format of the narrative. The model is instructed to avoid using tables or lists, and present the narrative as a coherent story. In addition, we instruct the model to aim at a narrative of 10 sentences, and involve only the 4 most important features (by absolute value) in the narrative, which is also precisely the size of the truncated SHAP table.

\begin{figure*}[!htbp]
    \includegraphics[width=0.95\textwidth, angle=0]{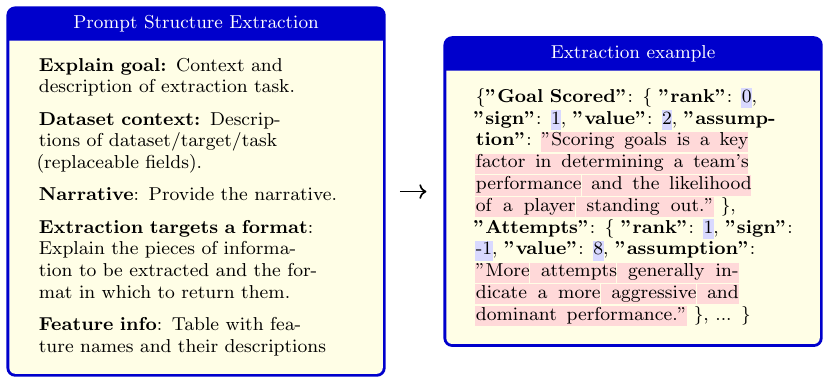}
    \caption{\textbf{Left:} A brief summary of the zero-shot prompt that we use to extract the narratives. \textbf{Right:} Excerpt from an extraction generated with gpt-4o for the narrative from Fig.~\ref{fig:GenerationPrompt} formatted as a dictionary. The same elements highlighted in the Fig.~\ref{fig:GenerationPrompt} can be found in the extracted dictionary in the same color code. }
    \label{fig:ExtractionPrompt}
\end{figure*}

\section{Methodology}\label{sec:methodology}

\textbf{Extractions:} We propose to use an extraction LLM that is instructed to extract information from a given narrative as visualized in Fig.~\ref{fig:ExtractionPrompt}, which can then be validated using downstream metrics. The set of all $N$ input features $\mathcal{F}$ of the underlying prediction model, is always passed to the extraction model as part of the prompt. The extraction model then extracts a dictionary with keys $f_{j} \in \mathcal{F} : j \in \lbrace{0,...,n-1 \rbrace}$ representing the features present in the narrative. Keep in mind that in our case the generation model receives a truncated table with only the most important features and hence $n < N$. For every feature $f_j$ identified in the narrative, the following quantities are extracted:

\begin{itemize}
    \item \textbf{Rank}: integer $ 0 \leq r_j \leq n-1$ that represents the importance rank of feature $f_j$ as implied in the narrative. In the present work we chose to use absolute-value based rankings and so $r=0$ represents the feature that is described as most significant regardless of its sign. Typically, the extraction order of the features follows the rank and hence $r_j=j$. 
    \item \textbf{Sign:} $s_j \in \lbrace{ -1,1\rbrace}$, represents whether feature $f_j$ was described to contribute positively or negatively to the score. To avoid the ambiguity of SHAP having an opposite sign for the two outcome classes, we explicitly instruct the generation model to formulate the feature contribution relative to class 1 and only provide the SHAP values for that class.
    \item \textbf{Value:} $v_j \in \mathbb{R} \cup \lbrace \phi \rbrace$ represents the actual value of the feature according to the narrative. Since we leave it up to the generation model which feature values to explicitly mention, we restrict the instruction to extract only the values that were mentioned in the narrative, and otherwise return a null-value $\phi$.
    \item \textbf{Assumption}: The extraction model extracts an assumption $a_j \in \mathcal{A} \cup \lbrace{ \phi \rbrace}$, which we instruct to be a single sentence. Similarly to values, if an assumption is not present in the narrative, we ask the model to extract a null-value $\phi$ (string 'None').
\end{itemize}

The faithfulness of the narrative is captured in the first three quantities of rank, sign and values. Although the feature values themselves are technically not part of SHAP, they are part of the LLM input and fall within our definition of faithfulness. The decision to consider these quantities is inspired from the rank-based and sign-based metrics proposed in the context of the disagreement problem \cite{krishna2024Disagreement}. On the other hand, the assumptions are more elusive and do not relate to the original data or model, and represent external knowledge injected by the generation model. 

\textbf{Metrics for (SHAP) faithfulness:} Validating the extracted information can be done with a range of downstream metrics of which we present possible examples. For SHAP faithfulness, the downstream metrics should measure Rank Agreement (RA), Sign Agreement (SA) and Value Agreement (VA) to the ground truth values $r^*_j, s^*_j, v^*_j$, where for example $r^*_j$ represents the actual rank of the extracted feature $f_j$. Metrics of this type have also been proposed in \cite{krishna2024Disagreement} and used in a recent related work using LLMs for XAI \cite{Kroeger2024}, although not for a narrative. Here, we will use the most straightforward version that simply measures the accuracy for every quantity (RA/SA/VA):
\begin{equation}
    XA = \sum_{x_j \neq \phi }  \frac{\delta_{x_j, x^*_j}}{n- \sum_{x_j = \phi} 1 }.
    \label{eq:XA}
\end{equation}
In the rare case that an extracted feature $f_j$ does not exist in $\mathcal{F}$, it is omitted from the accuracy calculation. For VA, we only count the values that were extracted and omit values $\phi$ for which no numerical quantity could be identified in the text. The latter is not necessarily unwanted behavior since we leave it up to the generation model which values to present in the narrative. Natural extensions could be to consider adding weights, or moving towards Kendall-Tau or permutation-based metrics for the rank.

\begin{figure*}[htbp!]
    \centering
    \includegraphics[width=0.95\textwidth]{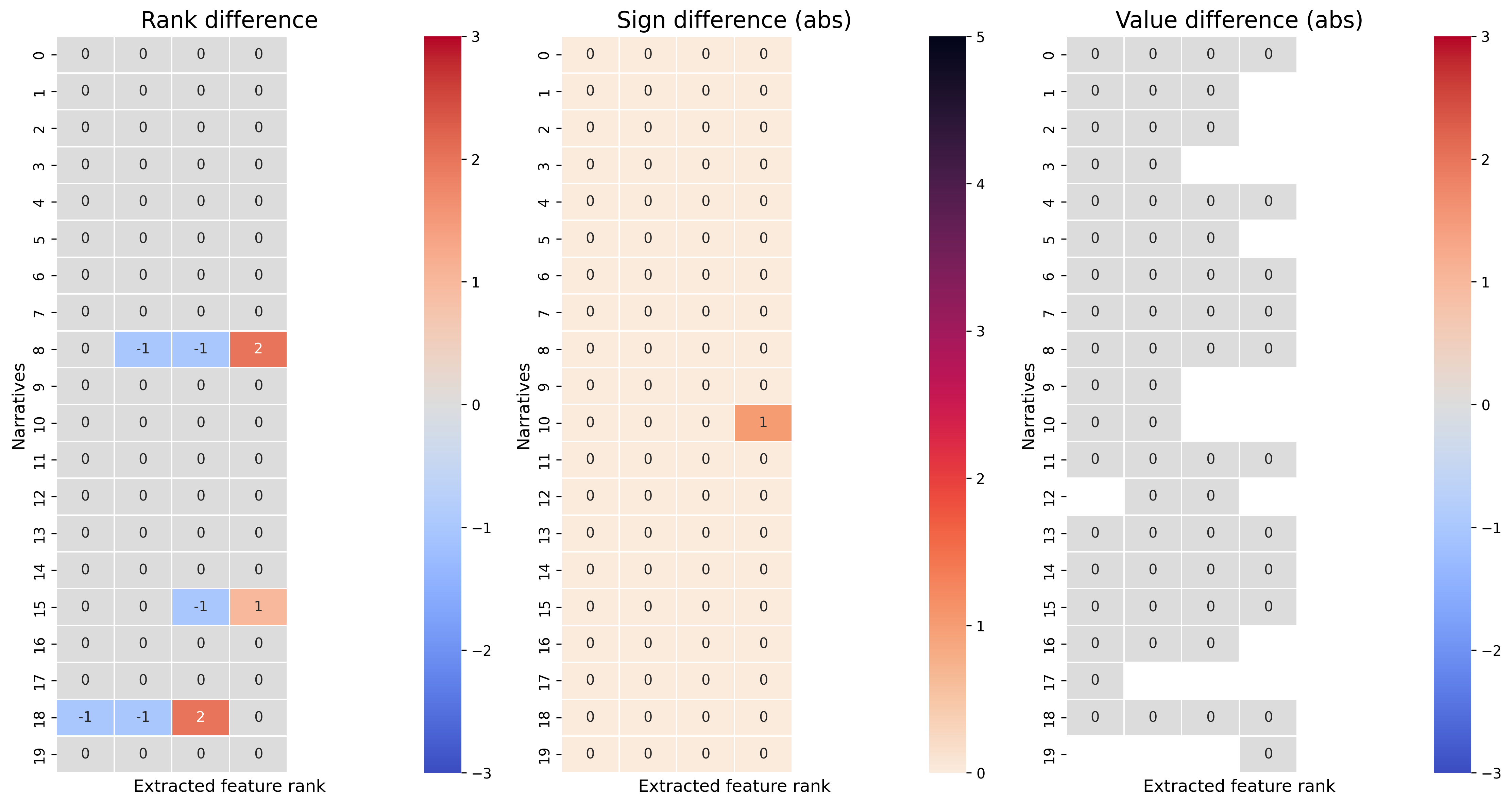}
    \caption{Overview tables for SHAP faithfulness, reflecting the differences between the extracted quantities and their ground truth for 20 narratives of the Fifa dataset generated with gpt-4o. Note that the absence of values on the most right panel indicates $\phi$ (meaning no value was identified in the narrative).}
    \label{fig:sf_example}
\end{figure*}

\textbf{Metrics for assumptions}: The assumptions of a narrative are a list of statements $a_j$ corresponding with each feature $f_j$, e.g. \textit{"Goal Scored": "Scoring goals is a key
factor in determining a team’s
performance and the likelihood
of a player standing out"}. One possible automated metric that can be used to assign a numerical value to a statement is \textit{perplexity}. Given a string of tokens $\lbrace{u_0,u_1 , ... u_{N-1} \rbrace}$, and the auto-regressive probability learned by a base LLM $p_i=p(u_i|u_{0}, u_1, ... , u_{i-1})$, perplexity is defined as:
\begin{equation}
PPL= \left(  \prod_{i=0}^{N-1} p_i \right)^{-1/N} .
\end{equation}
Perplexity measures how unlikely it is for the LLM to produce this string which as a rough proxy has been used for the purposes of fact-checking \cite{Lee2021, yuan2024probelm} or other evaluation tasks \cite{Wang2024}. In Sec.~\ref{sec:results} we will first validate this metric for our use case and show that perplexity is lower for more reasonable assumptions. Nevertheless, it should be emphasized that perplexity has also been criticized \cite{Wang2023} in the literature and hence should be used with care. 

One natural concern to have is whether anything of interest is added by first generating an assumption using an LLM and then measuring it relative to an LLM again. However, note that by extracting our assumptions from the narrative we can measure their perplexity as stand-alone statements. This can make a significant impact since it might have been the context of the narrative that induced an unreasonable assumption. For example, imagine that a SHAP table misleadingly suggests that exercise has a negative correlation with health outcomes, hereby forcing the generation LLM to search for interpretations. However, in vacuum the statement \textit{Exercise is unhealthy} has almost three times the perplexity of \textit{Exercise is healthy} (relative to Llama-3 8B) and hallucinations of this nature should on paper therefore be immediately captured with this metric (although we will see that this is more subtle in practice in Sec.~\ref{sec:results}).

\textbf{Metrics for human similarity:} As a complementary avenue that is not coupled to the extractions, we also explore the use of embeddings to compare the LLM generated narratives to human written narratives. The practical workflow to compare embeddings is quite straightforward. First, an embedding model transforms a string of text (in our case the entire narrative) into a high-dimensional vector. Then, for any two narratives embedded as vectors $\mathbf{a}$ and $\mathbf{b}$ with an angle $\theta_{ab}$ the similarity distance is simply defined as $d_s= 1 - \cos(\theta_{ab} )$. For early embedding approaches that created embedded representations at word level \cite{Mikolov2013}, such a distance would measure the semantic similarity of words across multiple dimensions. More recent techniques allow creating embeddings at a full sentence level and beyond \cite{Reimers2019SentenceBERTSE, Lee2024Embed}.

Embedding-based metrics such as BLEURT have already been used for this purpose in \cite{Burton2023} and will also be used further as a reference. However, here we want to demonstrate that even the simple cosine similarity metric used on embeddings created by state-of-the-art embedding models, already captures several interesting properties and can compare to BLEURT. This indicates that any modern embedding model can potentially be used as a solid basis for future fine-tuned narrative metrics.

\section{Experiments}\label{sec:results}

\subsection{Metrics validation}\label{subsection:mvalidation}

\textbf{Experiments setup:} For our experiments we use the same three binary classification datasets as in \cite{martens2023}: \textit{Fifa Man of the Match} (predicting man of the match winner), \textit{German Credit Score} (predicting good vs bad credit risk) and \textit{Student Performance Dataset} (predicting student passing or not). The target classification model to be explained is always a Random Forest (RF) with the default scikit hyperparameters. The SHAP values are generated with the SHAP package\footnote{https://github.com/shap/shap} using the Tree Explainer method. Per dataset, we select 20 instances (10 per ground truth class) for narrative generation. 

For the generation models we consider: gpt-4o \cite{openai2024gpt4technicalreport}, Claude Sonnet 3.5 \cite{ClaudeSonnet}, Llama-3 70b \cite{llama3modelcard,grattafiori2024llama3herdmodels}, Mistral-Large-2407 \cite{mistral2407}, and human-written (written by our research team). As an extraction model we always use gpt-4o, and demonstrate that it is highly accurate. For the LLMs we consider two types of prompts, a long one (the one explained in Sec.~\ref{sec:generation}), and a short one which still contains all data but receives more brief instructions. For the human-written narratives, the writers were instructed with the same long LLM prompt, but asked to avoid trivial identifiers. For all other details we refer the reader to our paper repository\footnote{\href{https://github.com/ADMAntwerp/SHAPnarrative-metrics}{https://github.com/ADMAntwerp/SHAPnarrative-metrics}}. 

\textbf{Extraction model:} Consider a classification task to determine whether a narrative is faithful to SHAP in rank and sign. The classifier compares the extracted quantities to the ground truth SHAP table and returns (1: all agree) or (0: at least one disagreement). In what follows we make rough estimates of the performance of the extraction model on this classification task to ensure it performs as expected. 

As a first test, we would like to confirm the reasonable expectation that the true negative (TN) rate is high and that if a narrative is not SHAP faithful, it will get flagged. This is indeed a property one would be interested in for practical applications. To do this we  generate a set of 60 manipulated narratives using the same generation workflow as described in Sec.~\ref{sec:generation}, with an additional manipulation step where we permute the order of the SHAP values among the features in the truncated SHAP table before passing it to the generation model. The goal is to create a sample of narratives with mistakes in them, and the specific choice of manipulation is not that important for our purpose. For reference, this is a weaker version of the manipulation step discussed later in Fig.~\ref{fig:ManipulatedVisualization}, with the main difference being that here the order swap is random and features are allowed to have the same sign after swap. Since undoing a random permutation should be rare, we will assume that all narratives generated in this way indeed have inaccuracies both in their rank and/or sign. We then pass all the manipulated narratives through an extraction model, and confirm that, as expected, after extraction all narratives were found to have errors in their rank and/or sign and would not have passed the classification. For both generation and extraction \textit{gpt-4o} was used. 

Similarly, to confirm that the false negative (FN) rate of this approach can also be expected to be low, we rely on our 60 human written narratives (20 per dataset). Since human narratives still contain some mistakes, we cannot use it in the same way as above and proceed with more care. We passed the narratives through the extraction model and found 8 out of 60 narratives to be flagged with a rank and/or sign error. 
Upon further inspection, 7 of these either contain full mistakes or ambiguous formulations from which a rank/sign cannot be determined, and only 1 was an actual false negative. Having already previously established that the false positive rate is very low, we assume that the remaining 52 narratives are then correct. Therefore, out of 53 correct narratives, only 1 was found to be a false negative.  

\begin{table}[h!]
    \centering
    \begin{tabular}{|c|c|c|c|}
        \hline
        \textbf{True Neg.} & \textbf{False Pos.} & \textbf{False Neg.} & \textbf{True Pos.} \\ 
        \hline
         60/60 &   0/60 &   1/53 &  52/53 \\
        \hline
    \end{tabular}
    \caption{Observed classification errors over the faithful (positive) and faulty (negative) narratives.}
\end{table}
We can conclude that the zero-shot prompted extraction model performs quite well and can generally be trusted, although more thorough experiments are needed to establish these rates more accurately. We proceed to use the extraction model on all the datasets and narratives, and generate extractions to measure SHAP faithfulness as depicted in Fig.~\ref{fig:sf_example}, of which the results will be discussed further in this section.

\begin{figure}[htbp!]
    \centering
\includegraphics[width=0.45\textwidth, angle=0]{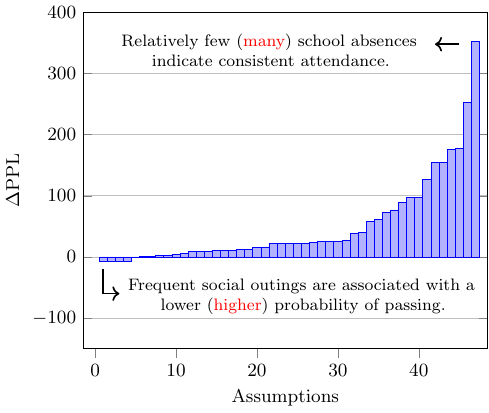}
\caption{The increase in perplexity after manipulating the assumptions (sorted). For illustration we also provide the assumption together with its manipulation (red) for the most negative and most positive perplexity change.}
\label{fig:PPLincrease}
\end{figure}

\textbf{Assumptions:} 
As described in Sec.~\ref{sec:methodology}, we also extract assumptions -- single sentences that summarize the implication or suggestion in the narrative as to why a particular feature could have contributed in a particular way. To validate the assumptions, we want to explore \textit{perplexity} as a downstream metric. Computing the perplexity of a reference sentence requires access to the token output probabilities of a model, which is why we will compute the perplexity relative to the \textit{Llama-3} base model with 8 billion parameters (later Mistral-7b will be used as well). We also apply weight quantization to reduce the memory requirements when loading these models for the perplexity calculation. To validate the use of perplexity for this purpose, we extract 50 randomly chosen assumptions from various narratives across datasets, and manually manipulate them to make them sound more unreasonable, with two examples of the manipulations given in Fig.~\ref{fig:PPLincrease}. We found that 3 out of 50 resembled factual statements related to the data (hence not actual assumptions) and proceed with the remaining 47. 

As can be seen on Fig.~\ref{fig:PPLincrease}, in this idealized setting, perplexity is indeed capable of detecting more unreasonable assumptions with good accuracy. The perplexity decreases only for five assumptions, and even then only by a small amount when compared to the average magnitude. Upon closer inspection, the assumption corresponding to the largest decrease (sentence at the bottom of Fig.~\ref{fig:PPLincrease}) is indeed more ambiguous than the one for the largest increase (sentence at the top of Fig.~\ref{fig:PPLincrease}). One can easily imagine arguing both ways as to why an active social life can contribute positively or negatively to the grades of a student, while having many absences almost by definition implies inconsistent attendance. 

It is important to emphasize that perplexity is sensitive to numerous other factors such as spelling or (un)common formulations unrelated to the assumptions of interest. Therefore, assumptions having an unexpectedly high perplexity in a particular narrative should warrant a closer inspection, but cannot be used as an exclusive measure. 

\textbf{Human similarity}:
We proceed to validate the last metric related to narrative embeddings as discussed in Sec.~\ref{sec:methodology}. Here, we want to gauge whether the distance between narratives measured with respect to a state-of-the-art text embedding can compare to BLEURT. In what follows \textit{voyage-large-2-instruct} \footnote{At the time of doing the experiments, this model held rank 1 on the Massive Text Embedding Benchmark (MTEB) Leaderboard.} will be used as the embedding model, which converts a narrative into a 1024-dimensional vector. For our purposes we limit ourselves to a simple cosine similarity metric. 

\begin{figure*}[htbp!]
    \centering
    \includegraphics[width=0.95\textwidth, angle=0]{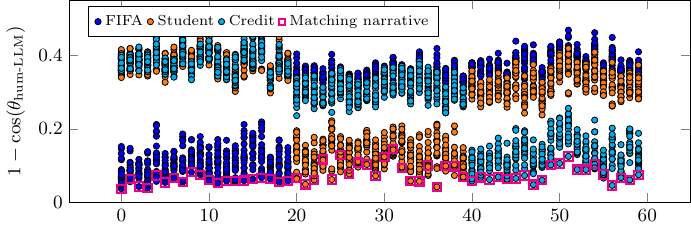}
    \includegraphics[width=0.95\textwidth, angle=0]{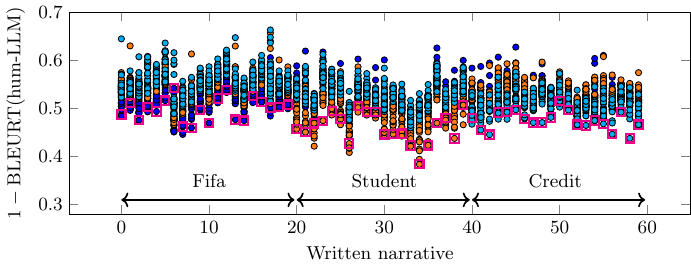}
    \caption{For every human written narrative out of 60 on the x-axis, the human similarity distance to 60 LLM-generated narratives is shown on the y-axis for the three datasets. The narrative corresponding to the same explanation is highlighted (magenta). Upper panel: cosine similarity, lower panel: BLEURT with BLEURT-20.}
    \label{fig:HumSimScatter}
\end{figure*}

On Fig.~\ref{fig:HumSimScatter} we present the results of a comparison between the 60 human written narratives and 60 narratives generated by gpt-4o on the same set of explanatizons and the same prompt\footnote{Humans were however explicitly asked to try and avoid trivial identifiers such as exact floating point numbers e.g. \textit{probability of 57.232\% to win the award.}} across our three datasets. The important observation is that in a majority of cases (42/60 for cosine similarity and 29/60 for BLEURT), the corresponding narrative generated from the same explanation provides the closest match. We see that, at least based from this discussion, the cosine similarity on a modern embedding outperforms BLEURT in the task of matching similar narratives. This motivates the use of a human similarity metric as the average embedding distance between a generated narrative to the corresponding human written narrative (i.e.~average value of the square magenta markers of Fig.~\ref{fig:HumSimScatter}). Of course, the cosine similarity merely provides a rough measure and it can be expected that the overwhelming change in $\theta$ arises due to similar themes and features mentioned in the narrative. Indeed, as we will see in Sec.~\ref{subsection:results}, while mismatches in rank and sign can be detected, they correspond to only a small part of the similarity. Therefore, the next natural step would be to fine-tune additional regression layers on top of the initial embedding to better isolate various properties of relevance. 

\subsection{Results}\label{subsection:results}

\textbf{Model Comparison:} Having established that the various metrics exhibit a reasonable behavior, we can now present a quantitative comparison between different LLMs and prompt types. For the purpose of reproducibility, in what follows we set the \textit{temperature} $T$ of both the generation and extraction LLMs to zero, which reduces the stochasticity of the narratives and extractions. Although for a practical setting one might arguably want more variation and creativity in the narrative generation, we did not notice any obvious drastic decrease in narrative quality at lower temperatures. We leave a more thorough investigation of the temperature effects to future studies. Also note that even at $T=0$ variations in output given the same prompt are still present to a varying degree among the models \cite{Ouyang}. To reduce this origin of stochasticity, further on we rerun identical experiments multiple times.

\begin{table}[htbp!]
    \centering
    \resizebox{0.49\textwidth}{!}{

% g-model,prompt,rank acc,sign acc,value acc,hum sim,bleurt,ppl_llama,ppl_mistral
% gpt-4o,long,0.925,0.975,0.961,,,,
% gpt-4o,short,0.796,0.896,0.965,,,,
% llama-3-70b,long,0.824,0.958,0.993,,,,
% llama-3-70b,short,0.637,0.768,0.943,,,,
% claude-3.5-sonnet,long,0.892,0.983,0.977,,,,
% claude-3.5-sonnet,short,0.755,0.903,0.967,,,,
% mistral-large-2407,long,0.896,0.988,0.988,,,,
% mistral-large-2407,short,0.714,0.924,0.972,,,,
% human,long,0.967,0.979,0.954,,,,

\large
    \setlength{\tabcolsep}{10pt}
    \begin{tabular}{ll|ccc}
          \toprule
          \textbf{Generation} & \textbf{Prompt} & \textbf{RA} & \textbf{SA} & \textbf{VA}  \\
          \midrule
          gpt-4o & long  & 0.925 & 0.975 & 0.961  \\
          gpt-4o & short & 0.796 & 0.896 & 0.965  \\
          llama-3-70b & long  & 0.824 & 0.958 & 0.993 \\
          llama-3-70b & short & 0.637 & 0.768 & 0.943   \\
          claude-3.5-sonnet & long  & 0.892 & 0.983 & 0.977   \\
          claude-3.5-sonnet & short & 0.755 & 0.903 & 0.967  \\
          mistral-large-2407 & long  & 0.896 & 0.988 & 0.988   \\
          mistral-large-2407 & short &  0.714 & 0.924 & 0.972  \\
  
      \end{tabular}}
    \caption{A comparison of the prompt types (long: the prompt with the various rules as discussed in Sec.~\ref{sec:generation}, short: minimal description of the task) for different generation models. The values for every row are averaged over 60 narratives across the three datasets.} 
    \label{table:ResultsTable}
\end{table}
To ensure that our prompt produces the intended results, we start by comparing our \textit{long} prompt with a \textit{short} minimalistic prompt, where both contain the necessary data and instruct the LLM to produce a narrative. The main difference is that the long prompt instructs the LLM in more detail what to pay attention to (e.g. rank and sign) and provides some other format and content related rules. The results are shown in Table~\ref{table:ResultsTable} for several popular models at the time of writing. Indeed, as would be expected from this setup, it can be clearly seen that for the faithfulness metrics the long prompt consistently achieves a better performance. This confirms that these instructions provide a measurable improvement and we can discard the \textit{short} prompt from consideration. 

\begin{table*}[htbp!]
    \centering
    \resizebox{0.98\textwidth}{!}{
 
\large
\setlength{\tabcolsep}{10pt}
\begin{tabular}{l|ccc|cc|cc}
    \toprule
    \multicolumn{1}{l|}{\textbf{Generation}} & \multicolumn{3}{c|}{\textbf{Faithfulness}} & \multicolumn{2}{c|}{\textbf{Human Similarity}} & \multicolumn{2}{c}{\textbf{Assumptions}} \\
    \cmidrule(r){1-8}  
    \textbf{Standard} & \textbf{RA} & \textbf{SA} & \textbf{VA} & $\mathbf{\cos(\theta)}$ & \textbf{BLEURT} &  \textbf{PPL} (L) & \textbf{PPL} (M) \\
    \midrule
      gpt-4o  &  0.904$\vert$0.975 & 0.983$\vert$0.987 & 0.962$\vert$0.987 & 0.926$\vert$0.930 & 0.522$\vert$0.526 & \hspace{6pt}95$\vert$103 & 71$\vert$77\\
      llama-3-70b  & 0.820$\vert$0.829 & 0.962$\vert$0.962 & 0.994$\vert$1.000 & 0.922$\vert$0.923 & 0.506$\vert$0.507 & 85$\vert$93 & 64$\vert$68\\
      claude-3.5-sonnet  & 0.879$\vert$0.908 & 0.983$\vert$0.988 & 0.982$\vert$0.984 & 0.923$\vert$0.925 & 0.512$\vert$0.515 & 84$\vert$88  & 62$\vert$68\\
      mistral-large-2407   & 0.838$\vert$0.867 & 0.983$\vert$0.996 & 0.981$\vert$0.989 & 0.925$\vert$0.928 & 0.510$\vert$0.513 & 74$\vert$85 & 57$\vert$67 \\
      human   & 0.936$\vert$0.966 & 0.971$\vert$0.975 & 0.933$\vert$0.954 & 1.000$\vert$1.000 & 0.929$\vert$0.929 & 85$\vert$92 & 65$\vert$72\\
      \midrule
      \textbf{Manipulated} & \textbf{RA} &  \textbf{SA} & \textbf{VA} &  $ \Delta \mathbf{\cos(\theta)}$ & $\Delta$\textbf{BLEURT} &  $\Delta$\textbf{PPL} (L) & $\Delta$\textbf{PPL} (M) \\
      \midrule
      gpt-4o & 0.012$\vert$0.037 & 0.288$\vert$0.312 & 0.978$\vert$0.993 & -0.0109$\vert$-0.0015 & -0.0221$\vert$-0.0159 & -7$\vert$13 & -4$\vert$15\\
      llama-3-70b & 0.054$\vert$0.062 & 0.308$\vert$0.312 & 0.979$\vert$0.979 & -0.0062$\vert$-0.0051 & -0.0097$\vert$-0.0089 & \hspace{-1pt}-18$\vert$-12  & \hspace{-8pt}-14$\vert$-9\\
      claude-3.5-sonnet & 0.008$\vert$0.025 & 0.237$\vert$0.267 & 0.976$\vert$0.987 & -0.0060$\vert$-0.0033 & -0.0229$\vert$-0.0192 & -1$\vert$14  & 0$\vert$9\\
      mistral-large-2407 & 0.017$\vert$0.037 & 0.150$\vert$0.175 & 0.980$\vert$0.994 & -0.0081$\vert$-0.0047 & -0.0248$\vert$-0.0177 &\hspace{-8pt}-16$\vert$-4& \hspace{-8pt}-16$\vert$-4\\
  \end{tabular}
 
}
    \caption{The results for the metrics discussed in this work across different generation models both for truthful generation (top part) and for misleading manipulated narratives (bottom part). Note that for the manipulated narratives, the last four columns display the observed relative increase in metric values over the top part. The values for every row are averaged over 60 narratives across the three datasets and rounded to the displayed decimals. To account for generation/extraction fluctuations, this is repeated 4 independent times for which the min$\vert$max values are reported. The perplexities were calculated with two smaller quantized models: Llama-3-8b (L) and Mistral-7B-v0.3 (M).} 
    \label{table:AveragedTable}
\end{table*}
Before discussing the results in more detail, note that the rank accuracy should be interpreted with some caution. The rank errors often arise from two feature ranks being swapped, which will then be counted as two errors when using Eq.~\ref{eq:XA}. Moreover, we emphasize that the errors in faithfulness might not only originate from actual mistakes in the narratives, but also from ambiguities when the rank or sign can simply not be deduced. This is not the case for the value agreement because in our approach we do not insist in having a value present in the narrative and the accuracy is computed only over values where the extraction model identified its presence. As discussed in Sec.~\ref{sec:methodology}, this is done to avoid having the narratives be filled with numeric values and improve reading fluency.

Having decided on the prompt, we present more elaborate results in Table~\ref{table:AveragedTable} which includes all metrics divided into the categories introduced in Sec.~\ref{sec:introduction}. The metrics are averaged over all narratives coming from all datasets, and more details can be found in Tables.~\ref{table:ResultsTableFifa}, \ref{table:ResultsTableStudent}, \ref{table:ResultsTableCredit} further on. In addition to this, we perform four identical runs and provide a range for the remaining fluctuations at zero temperature. We also add the metrics for human written narratives, which as discussed earlier can also contain mistakes and are not to be thought of as the sole golden standard.

For the standard narrative generation (top part of Table~\ref{table:AveragedTable}) we can observe that for the sign and value agreement, all generation models perform comparably well, while the differences for the rank are more appreciable. The human similarity metrics are computed on every narrative relative to the corresponding human written narrative for that same instance, which is also why these values drastically differ for the \textit{human} row. Being a trained metric, BLEURT exhibits some variation across identical string comparisons and hence does not yield exactly 1 there. We can observe a good degree of consistency where both of the similarity metrics are in agreement on the most and least similar models (not counting human), and for the perplexities we can even observe exactly the same ranking of the models across the two metrics. We remind the reader that the perplexities are measured on the extracted assumptions for every feature, rather than the entire narrative.

\begin{figure*}
    \includegraphics[width=0.96\textwidth, angle=0]{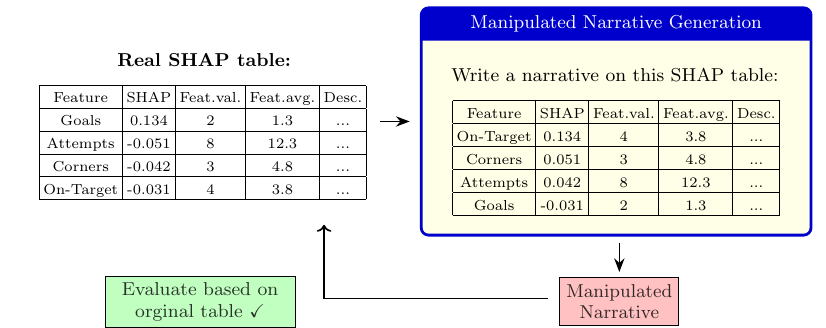}
    \caption{To create a manipulated narrative for the experiments discussed in Sec.~(\ref{subsection:results}) an LLM is prompted with a manipulated (truncated) SHAP table using the same prompt as for standard narrative generation (simplified here). In particular: (1) the absolute SHAP values stay in their place but the order of the most important features is inverted and (2) the sign of the SHAP values is changed to the opposite of the original sign for that feature. The feature values, averages and descriptions are kept together with the feature.}
    \label{fig:ManipulatedVisualization}
\end{figure*}

\textbf{Manipulated Narratives: } To provide a reference of how a misleading or faulty narrative generation would be detected on these metrics, we also present results for \textit{manipulated} narratives. This is visualized in Fig.~\ref{fig:ManipulatedVisualization}. We invert the rank order and signs of the features that are passed to the generation model at the prompt level, with all other evaluation steps proceeding as if the narrative was not manipulated.

Let us first go over all the results that are in line with the expectations. The rank accuracy is near zero and only rarely gets fixed by accident (e.g.~accidental swaps between feature 3 and 4). The value accuracy remains to be very good since the values were not manipulated in any way. Both the cosine similarity and BLEURT are also consistently in agreement on the fact that the manipulated narratives are further away from the human written narratives, although we can see that the differences are comparatively small. This is to be expected since the majority of the keywords remain the same and still result in a strong match, and indicates that training additional specialized layers on top of the embedding space could be of interest. 

However, a couple of more surprising results can also be observed. First, while one would expect the perplexities to increase due to more unreasonable assumptions for the swapped signs, this is not found across all models. To ensure that this is not simply an effect of the remaining fluctuations in generation and extraction, we have also confirmed that in each of the individual tables that were used to obtain the ranges in Table.~\ref{table:AveragedTable}, this decrease is also observed. One of factors contributing to this result is that compared to our idealized setup on Fig.~\ref{fig:PPLincrease} where we see a consistent increase, the LLMs are likely trying to make the text sound as plausible as possible and the resulting assumption is not simply a strict negation of the original one.

The most surprising result relates to the sign and we can see that depending on the model the sign accuracy reaches 30$\%$, while the naive expectation based on the top part of the table would be around 4$\%$. Put differently, the manipulated signs appear to get corrected a surprisingly large number of times. This indicates that the performance of the LLMs, at least when considering the sign agreement, is an order of magnitude worse when an unreasonable (manipulated) input is passed which contains nonsensical claims such as \textit{scoring five goals reduces the team's probability of winning the man of the match}. We hypothesize that prompts that demand an explanation of a sufficiently unreasonable feature table are more likely to make the model deviate from the task and make the narrative sound more plausible, which could be seen as a particular type of hallucination. This observation seems conceptually related to a well-known problem of \textit{parametric knowledge} in the literature where LLMs prefer to trust their own internal knowledge base rather than a context that contradicts it \cite{Longpre2021,xie2024adaptive}. There are some indications that this can be improved through better prompting \cite{zhou2023}. 

\begin{figure*}[htbp!]
    \centering
    \includegraphics[width=0.44\textwidth, angle=0]{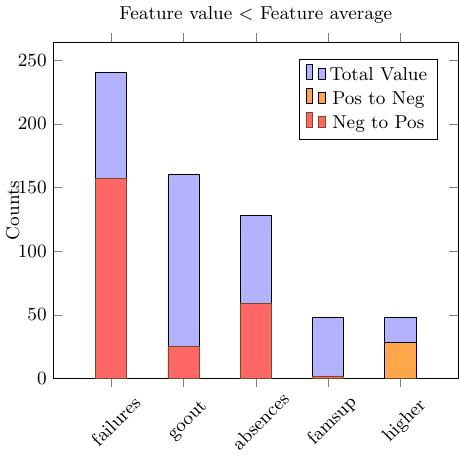}
    \includegraphics[width=0.4325\textwidth, angle=0]{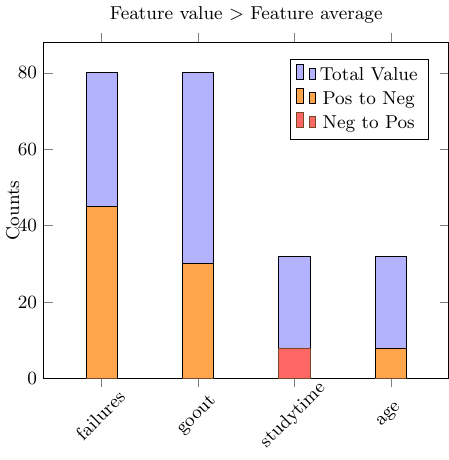}
    \caption{The counts of various types of feature swaps that happen relative to the manipulated SHAP table for the \textit{Student} dataset. We consider two cases where the feature values are smaller (left) and larger (right) than their average value in the table passed to the LLM. For reference the most important feature descriptions: (failures: number of failures in the past, gout: frequency of going out, absences: number of past absences).} 
    \label{fig:barplot_swaps}
\end{figure*}

Although this phenomenon requires further and more thorough research specifically for XAI narratives, we perform a first plausibility check where we look more closely at exactly which features are getting swapped in which direction. We consider the \textit{Student} dataset, and pool together all the manipulated narratives across the experiments performed to obtain Table~\ref{table:AveragedTable} (i.e. across different LLMs and noise averaging iterations). We then count the type of feature sign swaps (positive to negative or vice versa) over every feature present in this combined pool, and partition the counts in two feature classes respectively for the feature value being smaller or larger than the average value (which is also part of the input table to the generation model). To highlight the dominant part of the errors, we filter out all features that occur less than once in all 20 \textit{Student} instances (or less than 16 in the 320 pooled narratives). 

The results are presented on Fig.~\ref{fig:barplot_swaps}. It is clear that highest error rate (relative to the manipulated SHAP table) arises in the \textit{failures} feature which represents the number of class failures a student had in the past. Qualitatively, this is consistent with the hypothesis that the LLMs are self-correcting unreasonable statements. Indeed, looking through the other features it might be more difficult to explain more failures in the past contributing in a positive way compared to the rest. We can also see that the corrections happen in a direction that is more intuitive e.g.~going out too frequently is not good for your grades. It is important to highlight a subtlety in this interpretation. Note that the sign of the SHAP value itself is strongly correlated with whether the feature is larger or smaller than its average value, and the swap direction is therefore largely determined in advance. 

\begin{table*}[htbp!]
    \centering
    \resizebox{0.98\textwidth}{!}{ 
%%ORIGINAL VERSION WITH ALL METRICS
\large
\setlength{\tabcolsep}{10pt}
\begin{tabular}{l|ccc|cc|cc}
    \toprule
    \multicolumn{1}{l|}{\textbf{Student}} & \multicolumn{3}{c|}{\textbf{Faithfulness}} & \multicolumn{2}{c|}{\textbf{Human Similarity}} & \multicolumn{2}{c}{\textbf{Assumptions}} \\
    \cmidrule(r){1-8}  
    \textbf{Standard} & \textbf{RA} & \textbf{SA} & \textbf{VA} & $\mathbf{\cos(\theta)}$ & \textbf{BLEURT} &  \textbf{PPL} (L) & \textbf{PPL} (M) \\
    \midrule
      gpt-4o & 0.975 & 0.972 & 1.000 & 0.918 & 0.538 & 107.628 & 94.079\\
      llama-3-70b & 0.812 & 0.912 & 1.000 & 0.925 & 0.521 & 90.538 & 71.002\\
      claude-3.5-sonnet & 0.975 & 0.972 & 1.000 & 0.926 & 0.531 & 101.041 & 87.321\\
      mistral-large-2407 & 0.772 & 0.994 & 1.000 & 0.928 & 0.528 & 91.925 & 79.718\\
      human & 0.988 & 0.944 & 0.965 & 1.000 & 0.940 & 93.011 & 73.127\\
      \midrule
      \textbf{Manipulated} & \textbf{RA} &  \textbf{SA} & \textbf{VA} &  $ \Delta \mathbf{\cos(\theta)}$ & $\Delta$\textbf{BLEURT} &  $\Delta$\textbf{PPL} (L) & $\Delta$\textbf{PPL} (M) \\
      \midrule
      gpt-4o & 0.0156 & 0.3313 & 1.0000 & -0.0063 & -0.0178 & -3.8878 & -5.6022\\
      llama-3-70b & 0.0500 & 0.3188 & 1.0000 & -0.0065 & -0.0082 & -20.1002 & -16.7493\\
      claude-3.5-sonnet & 0.0281 & 0.3781 & 1.0000 & -0.0016 & -0.0188 & 3.0922 & -0.4203\\
      mistral-large-2407 & 0.0469 & 0.1375 & 1.0000 & -0.0052 & -0.0221 & -10.4193 & -19.2669\\ 
  \end{tabular}

 }    
    \caption{Similar to Table.~\ref{table:AveragedTable}, for only the \textbf{Fifa} dataset and averaged over the 4 iterations.} 
    \label{table:ResultsTableFifa}
\end{table*}

\begin{table*}[htbp!]
    \centering
    \resizebox{0.98\textwidth}{!}{ 
%%ORIGINAL VERSION WITH ALL METRICS
\large
\setlength{\tabcolsep}{10pt}
\begin{tabular}{l|ccc|cc|cc}
    \toprule
    \multicolumn{1}{l|}{\textbf{Student}} & \multicolumn{3}{c|}{\textbf{Faithfulness}} & \multicolumn{2}{c|}{\textbf{Human Similarity}} & \multicolumn{2}{c}{\textbf{Assumptions}} \\
    \cmidrule(r){1-8}  
    \textbf{Standard} & \textbf{RA} & \textbf{SA} & \textbf{VA} & $\mathbf{\cos(\theta)}$ & \textbf{BLEURT} &  \textbf{PPL} (L) & \textbf{PPL} (M) \\
    \midrule
      gpt-4o & 0.975 & 0.972 & 1.000 & 0.918 & 0.538 & 107.628 & 94.079\\
      llama-3-70b & 0.812 & 0.912 & 1.000 & 0.925 & 0.521 & 90.538 & 71.002\\
      claude-3.5-sonnet & 0.975 & 0.972 & 1.000 & 0.926 & 0.531 & 101.041 & 87.321\\
      mistral-large-2407 & 0.772 & 0.994 & 1.000 & 0.928 & 0.528 & 91.925 & 79.718\\
      human & 0.988 & 0.944 & 0.965 & 1.000 & 0.940 & 93.011 & 73.127\\
      \midrule
      \textbf{Manipulated} & \textbf{RA} &  \textbf{SA} & \textbf{VA} &  $ \Delta \mathbf{\cos(\theta)}$ & $\Delta$\textbf{BLEURT} &  $\Delta$\textbf{PPL} (L) & $\Delta$\textbf{PPL} (M) \\
      \midrule
      gpt-4o & 0.0156 & 0.3313 & 1.0000 & -0.0063 & -0.0178 & -3.8878 & -5.6022\\
      llama-3-70b & 0.0500 & 0.3188 & 1.0000 & -0.0065 & -0.0082 & -20.1002 & -16.7493\\
      claude-3.5-sonnet & 0.0281 & 0.3781 & 1.0000 & -0.0016 & -0.0188 & 3.0922 & -0.4203\\
      mistral-large-2407 & 0.0469 & 0.1375 & 1.0000 & -0.0052 & -0.0221 & -10.4193 & -19.2669\\ 
  \end{tabular}

 }    
    \caption{Similar to Table.~\ref{table:AveragedTable}, for only the \textbf{Student} dataset and averaged over the 4 iterations.}     
    \label{table:ResultsTableStudent}
\end{table*}

\begin{table*}[htbp!]
    \centering
    \resizebox{0.98\textwidth}{!}{ 

%%ORIGINAL VERSION WITH ALL METRICS
\large
\setlength{\tabcolsep}{10pt}
\begin{tabular}{l|ccc|cc|cc}
    \toprule
    \multicolumn{1}{l|}{\textbf{Credit}} & \multicolumn{3}{c|}{\textbf{Faithfulness}} & \multicolumn{2}{c|}{\textbf{Human Similarity}} & \multicolumn{2}{c}{\textbf{Assumptions}} \\
    \cmidrule(r){1-8}  
    \textbf{Standard} & \textbf{RA} & \textbf{SA} & \textbf{VA} & $\mathbf{\cos(\theta)}$ & \textbf{BLEURT} &  \textbf{PPL} (L) & \textbf{PPL} (M) \\
    \midrule
      gpt-4o & 0.941 & 1.000 & 0.927 & 0.925 & 0.530 & 111.997 & 74.261\\
      llama-3-70b & 0.812 & 0.988 & 0.991 & 0.912 & 0.503 & 112.791 & 83.169\\
      claude-3.5-sonnet & 0.863 & 0.984 & 0.949 & 0.911 & 0.517 & 99.914 & 63.467\\
      mistral-large-2407 & 0.922 & 0.988 & 0.960 & 0.917 & 0.518 & 85.983 & 55.382\\
      human & 0.963 & 1.000 & 0.914 & 1.000 & 0.936 & 91.160 & 67.786\\
      \midrule
      \textbf{Manipulated} & \textbf{RA} &  \textbf{SA} & \textbf{VA} &  $ \Delta \mathbf{\cos(\theta)}$ & $\Delta$\textbf{BLEURT} &  $\Delta$\textbf{PPL} (L) & $\Delta$\textbf{PPL} (M) \\
      \midrule
      gpt-4o & 0.0188 & 0.3594 & 0.9665 & -0.0063 & -0.0235 & 8.9043 & 8.7018\\
      llama-3-70b & 0.0875 & 0.4031 & 0.9375 & -0.0034 & -0.0037 & -12.5048 & -12.1190\\
      claude-3.5-sonnet & 0.0000 & 0.3375 & 0.9484 & -0.0076 & -0.0268 & -14.8936 & -4.8293\\
      mistral-large-2407 & 0.0187 & 0.2062 & 0.9570 & -0.0050 & -0.0226 & -11.2842 & -0.7269\\ 
  \end{tabular}

 }    
    \caption{Similar to Table.~\ref{table:AveragedTable}, for only the \textbf{Credit} dataset and averaged over the 4 iterations.} 
    \label{table:ResultsTableCredit}
\end{table*}
\section{Conclusion and outlook}
In this work we have explored several directions for metrics to evaluate XAI narratives on feature attribution explanations across three categories: Faithfulness, Human Similarity and Assumptions. After introducing and motivating the metrics, we perform several experiments to validate them and demonstrate that they can be expected to work. Next, we show how the metrics can be used to compare different LLMs for the narrative generation task and find roughly comparable performance to human-written narratives. To have a reference point on how the metrics perform on faulty or manipulated narratives, we also explore the evaluation of narratives where the explanation that was provided to the LLM is tinkered with. We find that in this scenario all metrics consistently decrease (except perplexity), indicating their utility in detecting faulty narratives. Surprisingly, we find that for the manipulated narratives the LLMs exhibit a stronger tendency to self-correct the sign of the feature contribution, providing an immediate use case for our metrics. We hypothesize that this could be a similar phenomenon to parametric knowledge bias where LLMs tend to correct context from the prompt that is in conflict with their internal knowledge base. This observation highlights a future challenge of LLM hallucinations for narratives on more subtle explanations.

Specifically for the various metrics we also identify the following interesting future directions (across the three categories):

\begin{itemize}
    \item \textbf{Faithfulness}: The metrics for faithfulness with the extraction-based pipeline work as expected and seem ready to use. However, they are by no means exhaustive, and it would be interesting to explore various additional metrics for this category.
    \item \textbf{Human Similarity}: We have demonstrated that modern embeddings contain sufficient resolution to match identical narratives (outperforming BLEURT even with a simple cosine similarity) and separate batches of unfaithful narratives from faithful ones. We believe that the most interesting next step would be to train metrics on top of the embedding model (or fine-tuning) that specialize in properties of XAI narratives. In particular, it seems relevant to move towards user preference and train automated metrics of narrative plausibility.  
    \item \textbf{Assumptions Plausibility}: We find that in an idealized setting for our validation experiments, perplexity relative to a smaller LLM is able to individually identify unreasonable assumptions. However, when applying this approach to the assumptions generated by LLMs in the manipulated narratives, the results are far less conclusive and perplexity does not appear to yield consistent behavior across models. Although this effect can be attributed to the LLMs trying to make the assumptions sound more plausible than in the idealized setting, it also indicates the need to improve assumption measurements beyond perplexity.
\end{itemize}

\section{Acknowledgments}

We acknowledge the support of the “Flemish AI Research Program” (FAIR), and the Research Foundation Flanders (FWO), Grant G0G2721N. 

\newpage
\twocolumngrid 
\bibliographystyle{unsrtnat}  
\bibliography{refs}

\begin{thebibliography}{39}
\providecommand{\natexlab}[1]{#1}
\providecommand{\url}[1]{\texttt{#1}}
\expandafter\ifx\csname urlstyle\endcsname\relax
  \providecommand{\doi}[1]{doi: #1}\else
  \providecommand{\doi}{doi: \begingroup \urlstyle{rm}\Url}\fi

\bibitem[Lundberg and Lee(2017)]{Lundberg2017}
Scott~M. Lundberg and Su-In Lee.
\newblock A unified approach to interpreting model predictions.
\newblock In \emph{Proceedings of the 31st International Conference on Neural Information Processing Systems}, NIPS'17, page 4768–4777, Red Hook, NY, USA, 2017. Curran Associates Inc.
\newblock ISBN 9781510860964.

\bibitem[Ribeiro et~al.(2016)Ribeiro, Singh, and Guestrin]{Tulio2016}
Marco~Tulio Ribeiro, Sameer Singh, and Carlos Guestrin.
\newblock "why should i trust you?": Explaining the predictions of any classifier.
\newblock In \emph{Proceedings of the 22nd ACM SIGKDD International Conference on Knowledge Discovery and Data Mining}, KDD '16, page 1135–1144, New York, NY, USA, 2016. Association for Computing Machinery.
\newblock ISBN 9781450342322.
\newblock \doi{10.1145/2939672.2939778}.
\newblock URL \url{https://doi.org/10.1145/2939672.2939778}.

\bibitem[Shapley et~al.(1953)]{shapley1953}
Lloyd~S Shapley et~al.
\newblock A value for n-person games.
\newblock 1953.

\bibitem[Huang and Marques-Silva(2023)]{huang2023}
Xuanxiang Huang and Joao Marques-Silva.
\newblock The inadequacy of shapley values for explainability, 2023.
\newblock URL \url{https://arxiv.org/abs/2302.08160}.

\bibitem[Marques-Silva and Huang(2024)]{silva2024}
Joao Marques-Silva and Xuanxiang Huang.
\newblock Explainability is not a game.
\newblock \emph{Commun. ACM}, 67\penalty0 (7):\penalty0 66–75, jul 2024.
\newblock ISSN 0001-0782.
\newblock \doi{10.1145/3635301}.
\newblock URL \url{https://doi.org/10.1145/3635301}.

\bibitem[Burton et~al.(2023)Burton, Al~Moubayed, and Enshaei]{Burton2023}
James Burton, Noura Al~Moubayed, and Amir Enshaei.
\newblock Natural language explanations for machine learning classification decisions.
\newblock In \emph{2023 International Joint Conference on Neural Networks (IJCNN)}, pages 1--9, 2023.
\newblock \doi{10.1109/IJCNN54540.2023.10191637}.

\bibitem[Martens et~al.(2023)Martens, Hinns, Dams, Vergouwen, and Evgeniou]{martens2023}
David Martens, James Hinns, Camille Dams, Mark Vergouwen, and Theodoros Evgeniou.
\newblock Tell me a story! narrative-driven xai with large language models, 2023.
\newblock URL \url{https://arxiv.org/abs/2309.17057}.

\bibitem[Zytek et~al.(2024)Zytek, Pidò, and Veeramachaneni]{zytek2024}
Alexandra Zytek, Sara Pidò, and Kalyan Veeramachaneni.
\newblock Llms for xai: Future directions for explaining explanations, 2024.
\newblock URL \url{https://arxiv.org/abs/2405.06064}.

\bibitem[Giorgi et~al.(2024)Giorgi, Campagnano, Silvestri, and Tolomei]{giorgi2024}
Flavio Giorgi, Cesare Campagnano, Fabrizio Silvestri, and Gabriele Tolomei.
\newblock Natural language counterfactual explanations for graphs using large language models, 2024.
\newblock URL \url{https://arxiv.org/abs/2410.09295}.

\bibitem[Pan et~al.(2024)Pan, Xiong, Wu, Zhang, Zhang, and Zhao]{pan2024}
Bo~Pan, Zhen Xiong, Guanchen Wu, Zheng Zhang, Yifei Zhang, and Liang Zhao.
\newblock Tagexplainer: Narrating graph explanations for text-attributed graph learning models, 2024.
\newblock URL \url{https://arxiv.org/abs/2410.15268}.

\bibitem[Cedro and Martens(2024)]{cedro2024}
Mateusz Cedro and David Martens.
\newblock Graphxain: Narratives to explain graph neural networks, 2024.
\newblock URL \url{https://arxiv.org/abs/2411.02540}.

\bibitem[Wojciechowski et~al.(2024)Wojciechowski, Lango, and Dusek]{Wojciechowski2024}
Adam Wojciechowski, Mateusz Lango, and Ondrej Dusek.
\newblock Faithful and plausible natural language explanations for image classification: A pipeline approach.
\newblock In Yaser Al-Onaizan, Mohit Bansal, and Yun-Nung Chen, editors, \emph{Findings of the Association for Computational Linguistics: EMNLP 2024}, pages 2340--2351, Miami, Florida, USA, November 2024. Association for Computational Linguistics.
\newblock \doi{10.18653/v1/2024.findings-emnlp.130}.
\newblock URL \url{https://aclanthology.org/2024.findings-emnlp.130}.

\bibitem[Kroeger et~al.(2024)Kroeger, Ley, Krishna, Agarwal, and Lakkaraju]{Kroeger2024}
Nicholas Kroeger, Dan Ley, Satyapriya Krishna, Chirag Agarwal, and Himabindu Lakkaraju.
\newblock In-context explainers: Harnessing llms for explaining black box models, 2024.
\newblock URL \url{https://arxiv.org/abs/2310.05797}.

\bibitem[Slack et~al.(2023)Slack, Krishna, Lakkaraju, and Singh]{Slack2023}
Dylan Slack, Satyapriya Krishna, Himabindu Lakkaraju, and Sameer Singh.
\newblock Explaining machine learning models with interactive natural language conversations using talktomodel.
\newblock \emph{Nature Machine Intelligence}, 5\penalty0 (8):\penalty0 873--883, Aug 2023.
\newblock ISSN 2522-5839.
\newblock \doi{10.1038/s42256-023-00692-8}.
\newblock URL \url{https://doi.org/10.1038/s42256-023-00692-8}.

\bibitem[Keane et~al.(2021)Keane, Kenny, Delaney, and Smyth]{ijcai2021p609}
Mark~T. Keane, Eoin~M. Kenny, Eoin Delaney, and Barry Smyth.
\newblock If only we had better counterfactual explanations: Five key deficits to rectify in the evaluation of counterfactual xai techniques.
\newblock In Zhi-Hua Zhou, editor, \emph{Proceedings of the Thirtieth International Joint Conference on Artificial Intelligence, {IJCAI-21}}, pages 4466--4474. International Joint Conferences on Artificial Intelligence Organization, 8 2021.
\newblock \doi{10.24963/ijcai.2021/609}.
\newblock URL \url{https://doi.org/10.24963/ijcai.2021/609}.
\newblock Survey Track.

\bibitem[Jacovi and Goldberg(2020)]{Jacovi2020}
Alon Jacovi and Yoav Goldberg.
\newblock Towards faithfully interpretable {NLP} systems: How should we define and evaluate faithfulness?
\newblock In Dan Jurafsky, Joyce Chai, Natalie Schluter, and Joel Tetreault, editors, \emph{Proceedings of the 58th Annual Meeting of the Association for Computational Linguistics}, pages 4198--4205, Online, July 2020. Association for Computational Linguistics.
\newblock \doi{10.18653/v1/2020.acl-main.386}.
\newblock URL \url{https://aclanthology.org/2020.acl-main.386}.

\bibitem[Sellam et~al.(2020)Sellam, Das, and Parikh]{SELLAM2020}
Thibault Sellam, Dipanjan Das, and Ankur Parikh.
\newblock {BLEURT}: Learning robust metrics for text generation.
\newblock In Dan Jurafsky, Joyce Chai, Natalie Schluter, and Joel Tetreault, editors, \emph{Proceedings of the 58th Annual Meeting of the Association for Computational Linguistics}, pages 7881--7892, Online, July 2020. Association for Computational Linguistics.
\newblock \doi{10.18653/v1/2020.acl-main.704}.
\newblock URL \url{https://aclanthology.org/2020.acl-main.704}.

\bibitem[Banerjee and Lavie(2005)]{Banerjee2005}
Satanjeev Banerjee and Alon Lavie.
\newblock {METEOR}: An automatic metric for {MT} evaluation with improved correlation with human judgments.
\newblock In Jade Goldstein, Alon Lavie, Chin-Yew Lin, and Clare Voss, editors, \emph{Proceedings of the {ACL} Workshop on Intrinsic and Extrinsic Evaluation Measures for Machine Translation and/or Summarization}, pages 65--72, Ann Arbor, Michigan, June 2005. Association for Computational Linguistics.
\newblock URL \url{https://aclanthology.org/W05-0909}.

\bibitem[Min et~al.(2023)Min, Krishna, Lyu, Lewis, Yih, Koh, Iyyer, Zettlemoyer, and Hajishirzi]{MIN2023}
Sewon Min, Kalpesh Krishna, Xinxi Lyu, Mike Lewis, Wen-tau Yih, Pang Koh, Mohit Iyyer, Luke Zettlemoyer, and Hannaneh Hajishirzi.
\newblock {FA}ct{S}core: Fine-grained atomic evaluation of factual precision in long form text generation.
\newblock In Houda Bouamor, Juan Pino, and Kalika Bali, editors, \emph{Proceedings of the 2023 Conference on Empirical Methods in Natural Language Processing}, pages 12076--12100, Singapore, December 2023. Association for Computational Linguistics.
\newblock \doi{10.18653/v1/2023.emnlp-main.741}.
\newblock URL \url{https://aclanthology.org/2023.emnlp-main.741}.

\bibitem[Chen et~al.(2024)Chen, Zhong, Ri, Zhao, He, Steinhardt, Yu, and Mckeown]{Chen2023}
Yanda Chen, Ruiqi Zhong, Narutatsu Ri, Chen Zhao, He~He, Jacob Steinhardt, Zhou Yu, and Kathleen Mckeown.
\newblock Do models explain themselves? {C}ounterfactual simulatability of natural language explanations.
\newblock In Ruslan Salakhutdinov, Zico Kolter, Katherine Heller, Adrian Weller, Nuria Oliver, Jonathan Scarlett, and Felix Berkenkamp, editors, \emph{Proceedings of the 41st International Conference on Machine Learning}, volume 235 of \emph{Proceedings of Machine Learning Research}, pages 7880--7904. PMLR, 21--27 Jul 2024.
\newblock URL \url{https://proceedings.mlr.press/v235/chen24bl.html}.

\bibitem[Ren and Liu(2023)]{REN2023}
Xuan Ren and Lingqiao Liu.
\newblock You can generate it again: Data-to-text generation with verification and correction prompting, 2023.
\newblock URL \url{https://arxiv.org/abs/2306.15933}.

\bibitem[Masterman et~al.(2024)Masterman, Besen, Sawtell, and Chao]{MASTERMAN2024}
Tula Masterman, Sandi Besen, Mason Sawtell, and Alex Chao.
\newblock The landscape of emerging ai agent architectures for reasoning, planning, and tool calling: A survey, 2024.
\newblock URL \url{https://arxiv.org/abs/2404.11584}.

\bibitem[Krishna et~al.(2024)Krishna, Han, Gu, Wu, Jabbari, and Lakkaraju]{krishna2024Disagreement}
Satyapriya Krishna, Tessa Han, Alex Gu, Steven Wu, Shahin Jabbari, and Himabindu Lakkaraju.
\newblock The disagreement problem in explainable machine learning: A practitioner{\textquoteright}s perspective.
\newblock \emph{Transactions on Machine Learning Research}, 2024.
\newblock ISSN 2835-8856.
\newblock URL \url{https://openreview.net/forum?id=jESY2WTZCe}.

\bibitem[Lee et~al.(2021)Lee, Bang, Madotto, and Fung]{Lee2021}
Nayeon Lee, Yejin Bang, Andrea Madotto, and Pascale Fung.
\newblock Towards few-shot fact-checking via perplexity.
\newblock In Kristina Toutanova, Anna Rumshisky, Luke Zettlemoyer, Dilek Hakkani-Tur, Iz~Beltagy, Steven Bethard, Ryan Cotterell, Tanmoy Chakraborty, and Yichao Zhou, editors, \emph{Proceedings of the 2021 Conference of the North American Chapter of the Association for Computational Linguistics: Human Language Technologies}, pages 1971--1981, Online, June 2021. Association for Computational Linguistics.
\newblock \doi{10.18653/v1/2021.naacl-main.158}.
\newblock URL \url{https://aclanthology.org/2021.naacl-main.158}.

\bibitem[Yuan et~al.(2024)Yuan, Chamoun, Aly, Whitehouse, and Vlachos]{yuan2024probelm}
Moy Yuan, Eric Chamoun, Rami Aly, Chenxi Whitehouse, and Andreas Vlachos.
\newblock {PR}ob{ELM}: Plausibility ranking evaluation for language models.
\newblock In \emph{First Conference on Language Modeling}, 2024.
\newblock URL \url{https://openreview.net/forum?id=k8KS9Ps71d}.

\bibitem[Wang et~al.(2024)Wang, Qiu, Yue, Guo, Zeng, Feng, and Shen]{Wang2024}
Yongjie Wang, Xiaoqi Qiu, Yu~Yue, Xu~Guo, Zhiwei Zeng, Yuhong Feng, and Zhiqi Shen.
\newblock A survey on natural language counterfactual generation, 2024.
\newblock URL \url{https://arxiv.org/abs/2407.03993}.

\bibitem[Wang et~al.(2023)Wang, Deng, Sun, and Meng]{Wang2023}
Yequan Wang, Jiawen Deng, Aixin Sun, and Xuying Meng.
\newblock Perplexity from plm is unreliable for evaluating text quality, 2023.
\newblock URL \url{https://arxiv.org/abs/2210.05892}.

\bibitem[Mikolov et~al.(2013)Mikolov, Chen, Corrado, and Dean]{Mikolov2013}
Tomas Mikolov, Kai Chen, Greg Corrado, and Jeffrey Dean.
\newblock Efficient estimation of word representations in vector space, 2013.
\newblock URL \url{https://arxiv.org/abs/1301.3781}.

\bibitem[Reimers and Gurevych(2019)]{Reimers2019SentenceBERTSE}
Nils Reimers and Iryna Gurevych.
\newblock Sentence-bert: Sentence embeddings using siamese bert-networks.
\newblock In \emph{Conference on Empirical Methods in Natural Language Processing}, 2019.
\newblock URL \url{https://api.semanticscholar.org/CorpusID:201646309}.

\bibitem[Lee et~al.(2024)Lee, Roy, Xu, Raiman, Shoeybi, Catanzaro, and Ping]{Lee2024Embed}
Chankyu Lee, Rajarshi Roy, Mengyao Xu, Jonathan Raiman, Mohammad Shoeybi, Bryan Catanzaro, and Wei Ping.
\newblock Nv-embed: Improved techniques for training llms as generalist embedding models, 2024.
\newblock URL \url{https://arxiv.org/abs/2405.17428}.

\bibitem[OpenAI et~al.(2024)OpenAI, Achiam, Adler, Agarwal, and et~al.]{openai2024gpt4technicalreport}
OpenAI, Josh Achiam, Steven Adler, Sandhini Agarwal, and et~al.
\newblock Gpt-4 technical report, 2024.
\newblock URL \url{https://arxiv.org/abs/2303.08774}.

\bibitem[Anthropic(2024)]{ClaudeSonnet}
Anthropic.
\newblock Claude sonnet 3.5.
\newblock 2024.
\newblock URL \url{https://www.anthropic.com/news/claude-3-5-sonnet}.

\bibitem[AI@Meta(2024)]{llama3modelcard}
AI@Meta.
\newblock Llama 3 model card.
\newblock 2024.
\newblock URL \url{https://github.com/meta-llama/llama3/blob/main/MODEL_CARD.md}.

\bibitem[Grattafiori et~al.(2024)Grattafiori, Dubey, Jauhri, and et~al.]{grattafiori2024llama3herdmodels}
Aaron Grattafiori, Abhimanyu Dubey, Abhinav Jauhri, and et~al.
\newblock The llama 3 herd of models, 2024.
\newblock URL \url{https://arxiv.org/abs/2407.21783}.

\bibitem[Team(2024)]{mistral2407}
The Mistral~AI Team.
\newblock Mistral large 2 2407.
\newblock 2024.
\newblock URL \url{https://huggingface.co/mistralai/Mistral-Large-Instruct-2407}.

\bibitem[Ouyang et~al.(2024)Ouyang, Zhang, Harman, and Wang]{Ouyang}
Shuyin Ouyang, Jie~M. Zhang, Mark Harman, and Meng Wang.
\newblock An empirical study of the non-determinism of chatgpt in code generation.
\newblock \emph{ACM Trans. Softw. Eng. Methodol.}, September 2024.
\newblock ISSN 1049-331X.
\newblock \doi{10.1145/3697010}.
\newblock URL \url{https://doi.org/10.1145/3697010}.
\newblock Just Accepted.

\bibitem[Longpre et~al.(2021)Longpre, Perisetla, Chen, Ramesh, DuBois, and Singh]{Longpre2021}
Shayne Longpre, Kartik Perisetla, Anthony Chen, Nikhil Ramesh, Chris DuBois, and Sameer Singh.
\newblock Entity-based knowledge conflicts in question answering.
\newblock In Marie-Francine Moens, Xuanjing Huang, Lucia Specia, and Scott Wen-tau Yih, editors, \emph{Proceedings of the 2021 Conference on Empirical Methods in Natural Language Processing}, pages 7052--7063, Online and Punta Cana, Dominican Republic, November 2021. Association for Computational Linguistics.
\newblock \doi{10.18653/v1/2021.emnlp-main.565}.
\newblock URL \url{https://aclanthology.org/2021.emnlp-main.565}.

\bibitem[Xie et~al.(2024)Xie, Zhang, Chen, Lou, and Su]{xie2024adaptive}
Jian Xie, Kai Zhang, Jiangjie Chen, Renze Lou, and Yu~Su.
\newblock Adaptive chameleon or stubborn sloth: Revealing the behavior of large language models in knowledge conflicts.
\newblock In \emph{The Twelfth International Conference on Learning Representations}, 2024.
\newblock URL \url{https://openreview.net/forum?id=auKAUJZMO6}.

\bibitem[Zhou et~al.(2023)Zhou, Zhang, Poon, and Chen]{zhou2023}
Wenxuan Zhou, Sheng Zhang, Hoifung Poon, and Muhao Chen.
\newblock Context-faithful prompting for large language models.
\newblock In Houda Bouamor, Juan Pino, and Kalika Bali, editors, \emph{Findings of the Association for Computational Linguistics: EMNLP 2023}, pages 14544--14556, Singapore, December 2023. Association for Computational Linguistics.
\newblock \doi{10.18653/v1/2023.findings-emnlp.968}.
\newblock URL \url{https://aclanthology.org/2023.findings-emnlp.968}.

\end{thebibliography}
\end{document}